\documentclass{article}

\usepackage[preprint]{neurips_2024}

\usepackage[utf8]{inputenc}
\usepackage[T1]{fontenc}
\usepackage{hyperref}
\usepackage{url}
\usepackage{booktabs}
\usepackage{amsfonts}
\usepackage{nicefrac}
\usepackage{microtype}
\usepackage{graphicx}
\usepackage{xcolor}

\title{A medical coding language model trained on clinical narratives from a population-wide cohort of 1.8 million patients}

\author{
  Joakim Edin\textsuperscript{1,2} \quad
  Sedrah Butt Balaganeshan\textsuperscript{1,3} \quad
  Annike Kj{\o}lby Kristensen\textsuperscript{1,4} \\
  \textbf{Lars Maal{\o}e\textsuperscript{2}} \quad
  \textbf{Ioannis Louloudis\textsuperscript{1,3}} \quad
  \textbf{S{\o}ren Brunak\textsuperscript{1,3}} \\[6pt]
  \textsuperscript{1}Section for Health Data Science and AI, Department of Public Health, University of Copenhagen, Denmark \\
  \textsuperscript{2}Corti, Copenhagen, Denmark \\
  \textsuperscript{3}Novo Nordisk Foundation Center for Protein Research, University of Copenhagen \\
  \textsuperscript{4}Danish Research Institute for Suicide Prevention, Mental Health Centre Copenhagen, Denmark \\[4pt]
  Correspondence: S{\o}ren Brunak, \texttt{soeren.brunak@sund.ku.dk}
}

\begin{document}

\maketitle

\begin{abstract}
\textbf{Background:} Medical coding summarizes clinical documentation into standardized codes essential for billing, research, and public health monitoring. However, manual coding is both time-consuming and prone to errors. While machine learning models have been proposed to automate this process, existing studies often rely on small, specialized datasets that poorly represent the real-world heterogeneity of patients, particularly in terms of multimorbidity.

\textbf{Methods:} We trained a language model to predict medical diagnostic codes using 5.8 million electronic health records from 1.8 million patients across all medical specialties (except adult psychiatry) in Eastern Denmark from 2006 to 2016. The model processes clinical notes, medications, and laboratory results to predict ICD-10 diagnosis codes. A test set of 270,000 patients was used to scrutinize performance across specialties and investigate discrepancies between human and model predictions.

\textbf{Results:} The model achieved an overall micro F1 score of 71.8\% and a top-10 recall of 95.5\%. Performance varied significantly by specialty, with F1 scores ranging from 91\% in clinical neurophysiology to 53\% in child and adolescent psychiatry. Specialties with well-defined standardized diagnostic criteria achieved higher F1 scores, whereas those dealing with diagnoses with a broader diagnostic spectrum or greater clinical ambiguity performed lower. Codes appearing predominantly as secondary diagnoses showed markedly lower F1 scores than primary diagnosis codes. We chose three predominantly secondary diagnoses with low F1 scores for further analysis (suicide-related behaviors, weight disorders, and hypertension). For these codes, the model identified thousands of cases that healthcare professionals had not coded. Manual validation of a subset of these model-identified cases showed that 76--86\% of the cases should have been coded. This validation suggests a failure to document clinically relevant secondary conditions, which contributed to the low F1 scores rather than model errors.

\textbf{Conclusions:} The model demonstrates strong performance, capable of automating coding for approximately 50\% of cases and providing accurate suggestions for most others. Importantly, our study suggests that there is systematic under-coding of secondary diagnoses in this dataset from Eastern Denmark. For the ICD-10 codes investigated in detail, the model successfully identified many missed diagnoses when appropriately calibrated. These findings have implications for the accuracy of epidemiological research, public health surveillance, and the understanding of complex comorbidities, commonly referred to as multimorbidity. While our results are specific to Eastern Denmark in this time-period, similar time constraints and reimbursement structures in other healthcare systems may suggest that this is a broader challenge. Automated coding systems may offer a practical solution to help healthcare professionals capture more of the secondary conditions without increasing the workload.
\end{abstract}

\section{Introduction}

The International Classification of Diseases (ICD), overseen by the World Health Organization (WHO), serves as the backbone of structured medical coding in modern healthcare. ICD codes provide a universal language for documenting medical conditions. These machine-readable codes enable critical healthcare functions: they ensure providers receive proper reimbursement, allow researchers to track disease trends, and help public health officials allocate resources effectively. However, the act of assigning medical codes, medical coding, is currently a slow and manual process. In Scotland, professional medical coders are reported to require roughly 7--8 minutes per case, while in the US, an average inpatient case has been reported to take more than thirty minutes~\citep{dongAutomatedClinicalCoding2022,tsengAdministrativeCostsAssociated2018,stanfillPreparingICD10CMPCS2014}. Beyond being time-consuming, medical coding is prone to errors. A systematic review of thirty-two studies estimated the median accuracy of primary diagnosis codes to be 80.3\% (IQR: 63.3--94.1\%)~\citep{burnsSystematicReviewDischarge2012}.

Medical coding errors can have serious consequences. Healthcare providers risk not being reimbursed, researchers may reach incorrect conclusions based on misleading data, and governments risk implementing suboptimal policies~\citep{InternationalStatisticalClassification2004}. As an example, the WHO reports suicide attempts to be systematically under-coded worldwide~\citep{worldhealthorganizationPreventingSuicideGlobal2014}. One consequence of this under-coding and underreporting is the inability to monitor and prevent suicide epidemics. This was evident in the early 2000s, when carbon monoxide poisoning from charcoal burning rapidly became a standard suicide method across Hong Kong, Taiwan, Japan, Korea, and Singapore~\citep{changRegionalChangesCharcoalBurning2014}. Today, we understand that restricting access to suicide means can be an effective prevention strategy~\citep{mannSuicidePreventionStrategies2005,hawtonLongTermEffect2013}. However, for the epidemic in the 2000s, such preventive measures could not be implemented due to the lack of reliable data~\citep{changRegionalChangesCharcoalBurning2014}. Similarly, awareness campaigns, which could have been beneficial~\citep{torokSystematicReviewMass2017,matsubayashiEffectPublicAwareness2014}, were not employed during the suicide epidemic.

To reduce errors and time spent on medical coding, recent studies have proposed machine learning models for analyzing clinical narratives and predicting appropriate ICD codes~\citep{dongAutomatedClinicalCoding2022,edinAutomatedMedicalCoding2023a}. These models can assist during the coding process by suggesting a list of ICD codes, from which a human coder can select the relevant codes faster. However, existing research suffers from major setbacks, typically being evaluated on small, highly specialized datasets that poorly represent the diversity and complexity of the real-world patient populations~\citep{dongAutomatedClinicalCoding2022,edinAutomatedMedicalCoding2023a}.

The MIMIC-III dataset has been widely used in medical coding research~\citep{jiUnifiedReviewDeep2023}. However, with 41,126 patients from one American Intensive Care Unit (ICU), it represents a narrow slice of the general population~\citep{johnsonMIMICIIIFreelyAccessible2016}. MIMIC-IV has expanded this to 163,368 patients, but still only covers one ICU and one emergency department from a single hospital, resulting in a less diverse group of patients~\citep{johnsonMIMICIVFreelyAccessible2023}.

We trained and evaluated a machine-learning model on 5.8 million electronic health records (EHRs) from 1.8 million Danish patients across all specialties (except adult psychiatry) in Eastern Denmark (covering Region Zealand and the Capital Region of Denmark, approximately 50\% of the population) from 2006 to 2016. The model processes clinical notes, medications, and lab results, and returns a confidence score between zero and one for each ICD-10 code (the tenth revision of ICD). The higher the score, the more confident the model is of the corresponding code. The unprecedented scale of this dataset allowed us to demonstrate how model performance scales with more training data and to identify which medical specialties pose the greatest challenges for automatic medical code prediction. However, we also uncovered a critical problem: human coders in our data systematically undercode secondary diagnoses. Codes that appear predominantly as secondary diagnoses had worse model performance. Moreover, when we investigated specific secondary diagnoses (obesity, hypertension, and suicide attempts) with which the model often disagreed with human coders, manual review showed that the model was correct 66--72\% of the time, indicating that human coders had incorrectly omitted codes. The correct secondary diagnosis codes appeared among the model's top 10 predictions in 90\% of the cases. With such a system, healthcare professionals could quickly select relevant codes from a short list, capturing more complete clinical information without additional time investment.

\section{Results}

\subsection{Large-scale data improves automated medical coding performance}

Our original data set comprised 2.6 million patients, which after pre-processing steps yielded a final data set of 2.075 million patients. The filtering mainly removed records with very short or extremely long narratives (see Methods). We split these data into training, validation, and test sets comprising 1.8 million, 5,000, and 270,000 patients, respectively (see Methods section). To evaluate the impact of training set size on automated medical coding performance, we trained a Pre-trained Language Model with Cross-Attention (PLM-CA) on progressively larger datasets, ranging from 50,000 to 1.8 million patients~\citep{edinUnsupervisedApproachAchieve2024a}. ICD-10 uses a tree-like five-level hierarchy, where the third level typically represents the disease category while higher levels further specify the patient's manifestation of the disease. We assessed performance by comparing model predictions with the ICD-10 codes at the third and fifth level annotated by physicians and secretaries at the hospitals.

Model-human agreement improved consistently with increasing training data size across all metrics (Table~\ref{tab:performance}). We measured F1 scores, which balance precision (the proportion of predicted codes that were correct) and recall (the proportion of correct codes that were found), providing a single performance metric between 0\% and 100\%. F1-micro scores, which give more weight to frequently occurring codes, increased from 59.9\% with 50,000 patients to 71.8\% with 1.8 million patients used for training. F1-macro scores, which weight all codes equally regardless of frequency, showed even larger improvement from 25.6\% to 47.3\%. This substantial improvement in F1-macro scores indicates that larger training datasets comprising patients with a high level of multimorbidity notably enhanced the model's ability to predict rare diseases and conditions, which are weighted higher in macro- than micro-averaged metrics.

For the model trained on the largest dataset, the exact match ratio reached 54.6\%, meaning the model and human agreed on all codes in over half of the cases. Notably, Recall@10 reached 95.5\%, indicating that when presenting the model's top 10 code suggestions, 95.5\% of human-assigned codes appeared within these recommendations. This high recall suggests significant potential for clinical efficiency: presenting physicians with the model's top 10 predictions could be substantially more time-efficient than searching through the entire set of possible codes, as the correct answer often appears among these suggestions.

We also evaluated the model on level-5 codes, the highest level in the ICD-10 hierarchy. With ten times more codes than level-3, resulting in more infrequent codes and ambiguous coding decisions, performance decreased as expected (Table~\ref{tab:performance}). However, even at this challenging granularity, Recall@10 remained strong at 86.0\%, demonstrating that the model could still capture most human-annotated codes within its top suggestions. However, for brevity, we focus on level-3 predictions in the rest of the paper.

\begin{table}[t]
\caption{Model performance when trained on progressively larger data sets. All models are evaluated on the same test set using the human-annotated codes as the ground-truth labels.}
\label{tab:performance}
\centering
\small
\begin{tabular}{@{}lccccccc@{}}
\toprule
& \multicolumn{6}{c}{\textbf{ICD-10 Level 3}} & \textbf{Level 5} \\
\cmidrule(lr){2-7} \cmidrule(l){8-8}
\textbf{Training data} & 50K & 100K & 200K & 500K & 1M & 1.8M & 1M \\
\midrule
F1 Micro & 59.9\% & 61.7\% & 65.8\% & 69.0\% & 70.1\% & 71.8\% & 54.6\% \\
F1 Macro & 25.6\% & 26.3\% & 30.5\% & 41.0\% & 43.6\% & 47.3\% & 10.1\% \\
Exact Match Ratio & 37.0\% & 42.5\% & 47.3\% & 51.7\% & 52.0\% & 54.6\% & 33.2\% \\
Recall@5 & 82.2\% & 85.3\% & 87.8\% & 89.8\% & 90.5\% & 91.4\% & 78.4\% \\
Recall@10 & 88.7\% & 90.9\% & 93.0\% & 94.4\% & 94.9\% & 95.5\% & 86.0\% \\
Recall@15 & 91.3\% & 93.1\% & 94.8\% & 96.0\% & 96.4\% & 96.9\% & 89.2\% \\
Mean Avg.\ Precision & 72.4\% & 74.9\% & 78.0\% & 80.7\% & 81.5\% & 82.8\% & 66.5\% \\
Precision@Recall & 62.9\% & 65.4\% & 69.0\% & 72.2\% & 73.1\% & 74.8\% & 56.1\% \\
\bottomrule
\end{tabular}
\end{table}

\subsection{Model-human agreement varies across medical specialties}

\begin{figure}[t]
\centering
\includegraphics[width=\textwidth]{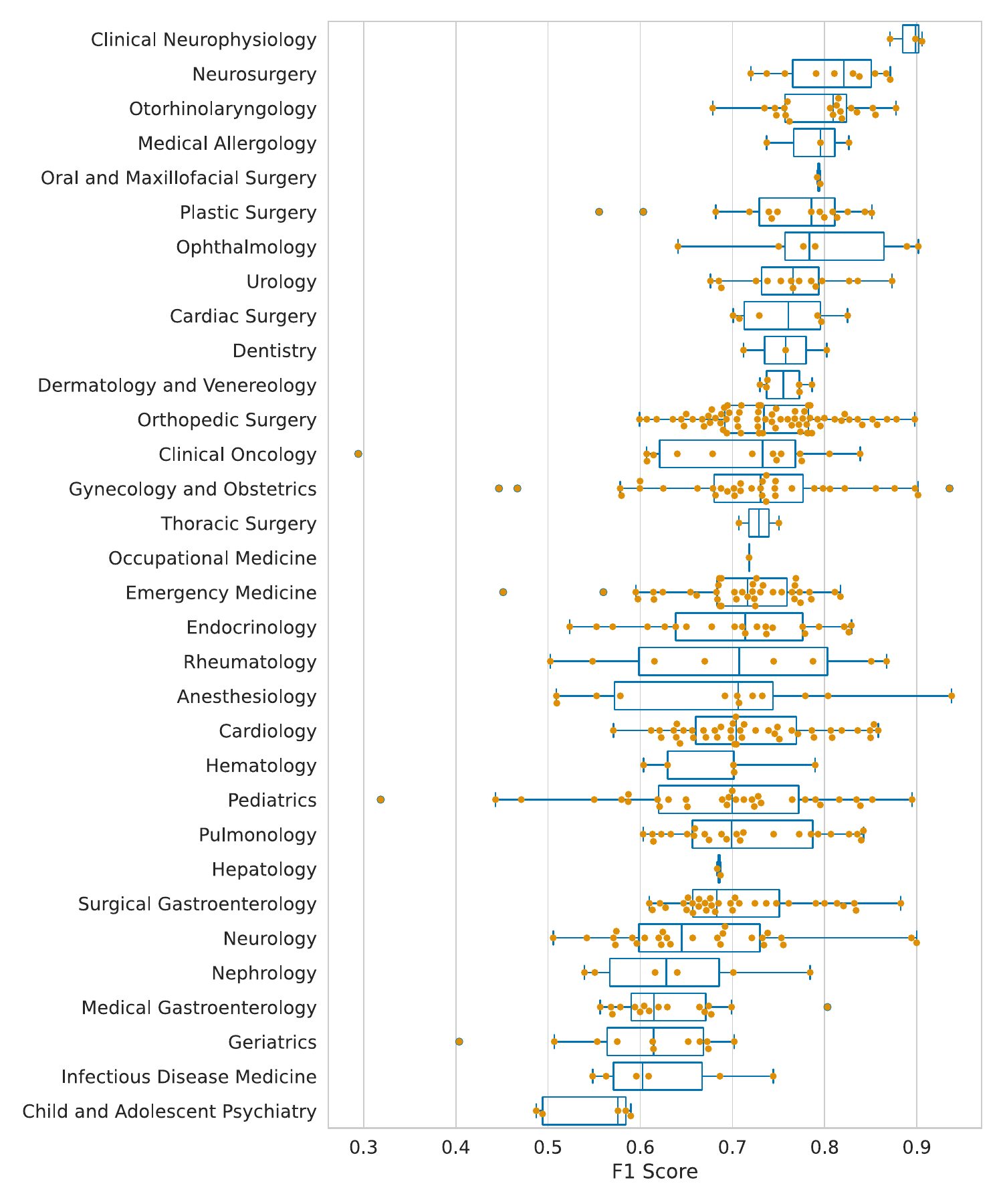}
\caption{The model's F1 score for each specialty. Each dot is the model's median F1 score for a specific department. The figure reveals a pattern in which the top specialties often involve planned admissions, while the bottom specialties represent patients with an average of many comorbidities. Departments with more than one hundred examples in the test set are included only.}
\label{fig:specialty}
\end{figure}

Figure~\ref{fig:specialty} presents F1-micro scores representing human-model agreement across medical specialties, with each data point showing the median score for a specific department. We excluded departments with fewer than 100 observations in the test set of 270,000 patients. The scores exhibit substantial variation, ranging from 53\% for child and adolescent psychiatry to 91\% for clinical neurophysiology.

Specialties with the highest F1 scores share common characteristics, which we assume are associated with more standardized workflows and clearer diagnostic pathways. Clinical neurophysiology exemplifies this pattern, as patients typically arrive with suspected diagnoses that healthcare professionals systematically confirm or reject through structured testing protocols. Many surgical specialties similarly achieve high scores because patients often present with already established diagnoses, making the coding process more straightforward for both humans and models. This predictability in workflow and diagnostic clarity translates directly into higher agreement rates between human coders and our model.

In contrast, specialties characterized by a higher degree of diagnostic uncertainty, numerous comorbidities, or less structured clinical workflows demonstrate notably lower agreement rates. Child and adolescent psychiatry shows the lowest score, which likely reflects multiple diagnostic complexities: parental reports that the parents' own perspectives may influence~\citep{ryanDiagnosingPediatricDepression2001}, children's difficulties in articulating their inner experiences~\citep{ryanClinicalPictureMajor1987}, and the fact that children typically present with multiple comorbidities by the time they reach psychiatric services~\citep{leafMentalHealthService1996,zuckerbrotImprovingRecognitionAdolescent2006}. Infectious disease diagnoses often remain uncertain throughout much of the admission period due to pending microbiology and/or other paraclinical results. This uncertainty creates particular coding difficulties because ICD-10 guidelines specify that uncertain diagnoses should be coded for inpatient cases but not outpatient cases~\citep{InternationalStatisticalClassification2004}. Since our model lacked information about the admission type, it was unable to account for this distinction. Moreover, the ICD-10 guidelines state that uncertain diagnoses should be coded for inpatient cases only, which physicians and secretaries may not consistently do, potentially creating inconsistencies in the training data. However, our large-scale retrospective analysis cannot assess how consistently these guidelines were followed in practice.

Specialties that commonly treat patients with multimorbidity also show consistently lower F1 scores. Geriatrics, nephrology, and neurology all fall into this category. Patients in nephrology often have morbidities that can cause kidney disease, such as diabetes, cardiovascular disease, and hypertension~\citep{macraeComorbidityChronicKidney2021}. Neurology faces similar complexity, as common neurological conditions such as stroke and Alzheimer's disease typically present alongside multiple comorbidities that complicate the coding process~\citep{santiagoImpactDiseaseComorbidities2021,cipollaImportanceComorbiditiesIschemic2018}. These cases require human coders to document numerous secondary diagnoses in addition to the primary reason for admission. As shown later, human coders often fail to code the patients' secondary diagnoses, contributing to the lower agreement rates observed in these complex specialties.

\subsection{ICD-10 codes with frequent human-model disagreements}

\begin{figure}[t]
\centering
\includegraphics[width=\textwidth]{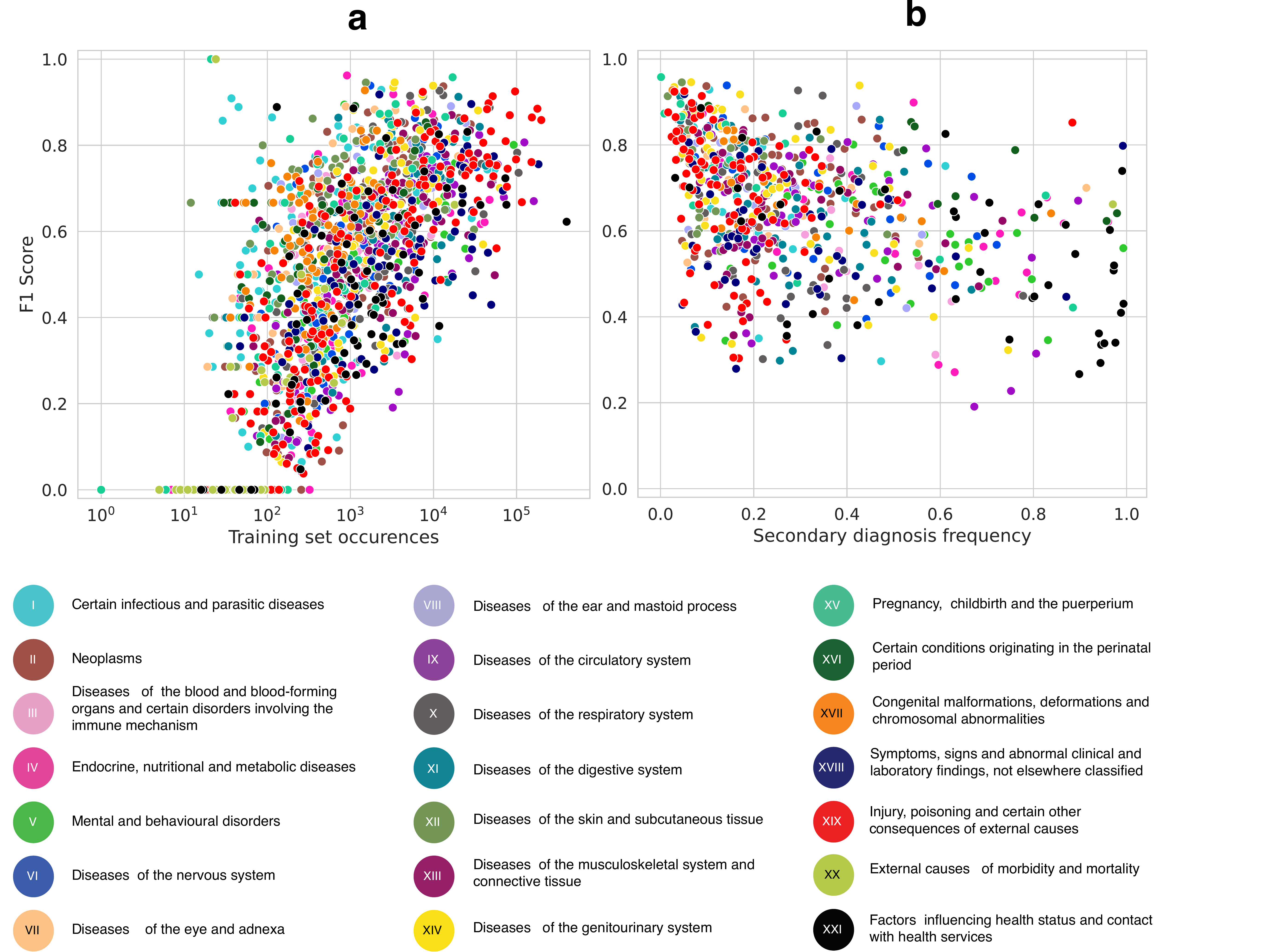}
\caption{a) The F1 score (y-axis) versus the number of occurrences in the training data (x-axis) for each ICD-10 code (level-3). b) The frequency of codes for secondary diagnoses in the training data (x-axis) and their F1 score (y-axis).}
\label{fig:code_characteristics}
\end{figure}

Figure~\ref{fig:code_characteristics} displays the relationship between code characteristics and human-model agreement at the ICD-10 level-3 category level. Figure~\ref{fig:code_characteristics}a depicts a strong inverse relationship between the mean F1 scores and the code frequency in the training dataset. The model never makes predictions for 20.1\% of all codes. These unpredicted codes predominantly occur at low frequencies in the training data. The logarithmic scale reveals that codes appearing fewer than 100 times in the training set show notably inferior performance, while codes with over 10,000 occurrences generally achieve mean F1 scores above 0.6.

\begin{figure}[t]
\centering
\includegraphics[width=\textwidth]{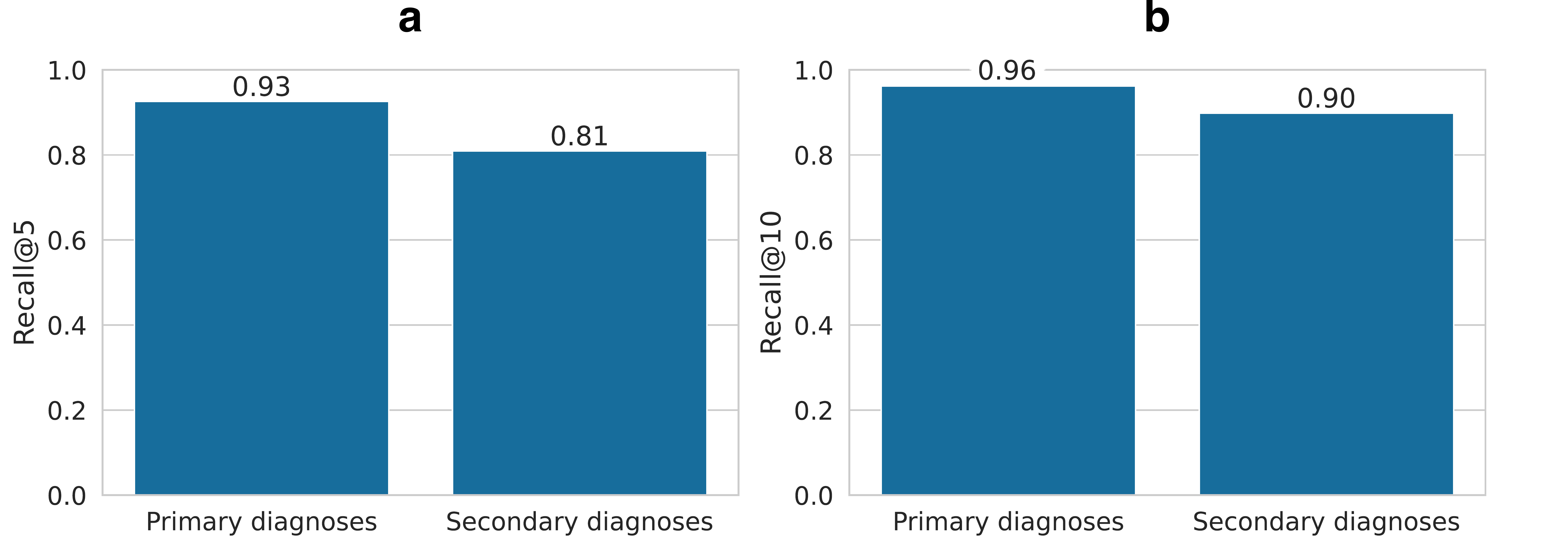}
\caption{a) Recall@5 for primary diagnoses and secondary diagnoses, and b) Recall@10. Recall@5 and Recall@10 measure the frequency of human-annotated codes among the model's top 5 and top 10 predictions.}
\label{fig:recall}
\end{figure}

However, code frequency provides only a partial explanation for performance variation. Notable exceptions exist where rare codes achieve high F1 scores and common codes show poor performance, indicating that factors beyond prevalence influence coding accuracy. Figure~\ref{fig:code_characteristics}b identifies one such critical factor: the role of codes as primary versus secondary diagnoses. Among codes with more than 1,000 occurrences in the training data, those that predominantly appear as secondary diagnoses show substantially lower mean F1 scores compared to those typically coded as primary diagnoses. When codes appear as secondary diagnoses in more than 80\% of cases, their mean F1 scores fall below 0.5, whereas codes primarily used as primary diagnoses maintain mean scores above 0.7. Furthermore, Figure~\ref{fig:recall} shows that the human-annotated primary diagnoses are more frequently among the model's top 5 and top 10 predictions than secondary diagnoses (Recall@5 and Recall@10). These patterns suggest that secondary diagnoses present unique challenges for both human coders and automated systems, warranting deeper investigation into the specific mechanisms driving these disagreements.

\subsection{What causes frequent human-model disagreements on secondary diagnoses?}

To understand the mechanisms behind the human-model disagreements on secondary diagnoses, we conducted detailed case studies of three conditions that exemplify different aspects of secondary diagnosis coding: Suicide-related behaviors (X60--X85), weight disorders (E66), and hypertension (I10--I15). For each condition, we experimented with decision thresholds and manually validated results to determine whether high model-human disagreement could stem from missing human annotations.

\subsubsection{Suicide-related behaviors}

Conventional measures of suicide-related behaviors (such as self-harm and suicide attempts) involve the ICD codes X60--X85. WHO defines the difference between self-harm and a suicide attempt as the intention to die. As intent is hard to extract reliably, these patient groups are often grouped as patients exhibiting suicide-related behaviours~\citep{reutermorthorstIncidenceRatesDeliberate2016}. The X60--X85 codes are underreported in Denmark, following a worldwide trend~\citep{worldhealthorganizationPreventingSuicideGlobal2014}. Among other things, the WHO cites the doctor's desire for their patient to avoid stigmatization as one of the reasons for the underreporting. Reuter et al.\ (2016) point to a pattern suggesting that time and resources might be to blame~\citep{reutermorthorstIncidenceRatesDeliberate2016}. Finally, it is worth noting that the codes that fall within the XX chapter are excluded from reimbursement calculations in Denmark, leaving little financial incentive for the code to be registered.

In the test set, 389 cases had been conventionally annotated by hand with reported suicide-related behaviors. At a standard decision boundary, the model detected only 21.3\% of these cases. When AttInGrad, a machine learning explanation method, was employed, we detected a pattern: the model correctly identified and relied on relevant textual features---such as explicit suicide mentions and used methods---yet paradoxically assigned these cases a low confidence score. This suggests the model learned to suppress predictions for suicide-related behaviors despite recognizing the relevant clinical evidence. We found that missing annotations in the training data were the most likely explanation.

Following previous studies, we also found that suicide-related behaviors were frequently not coded. Lowering the decision boundary to 0.05 improved detection to 64.5\% and uncovered 917 additional potential suicide-related events, which had not been coded by human hand. To validate these findings, we randomly selected 50 model-identified cases that lacked a human-coded X60--X85 code. Using AttInGrad to examine the model's decision-making features, we found 68\% of the EHRs to explicitly mention ``suicide attempt,'' 8\% to mention suicide related behaviors, and 12\% the cases described events which are often seen in suicide attempts---such as ingesting a high amount of paracetamol or lacerating the wrists with razor blades. We assumed those 12\% to be genuine suicide attempts and estimated 78\% to be genuine suicide attempts. Additionally, 8\% were self-harm---in total, we deemed 86\% of the model predictions to be correct. In the remaining 14\% of false positives, ambiguous overdoses, mentions of previous suicide attempts, or misinterpretations of text explicitly stating the patient had not attempted suicide were found.

\subsubsection{Weight disorders}

We found a similar pattern for weight disorders: overweight and obesity (E66). Weight disorders impact treatment protocols, medication dosing, and surgical risk assessment, but often constitute a secondary diagnosis (86.9\% of the cases in our dataset)~\citep{limObesityComorbidConditions2025}. The model detected 52.8\% of 4,870 human-annotated weight disorder cases at the standard decision boundary, increasing to 78.6\% when the threshold was set to 0.1. The model identified 6,551 potential weight disorder cases beyond human annotations. Manual validation of 50 randomly sampled model-identified cases revealed explicit mentions of overweight or high BMI in 76\% of the cases. Common errors arose from mentions of weight loss or gastric bypass procedures without confirmation of the current weight status.

\subsubsection{Hypertension}

Hypertension (I10--I15) covers a range of hypertensive diseases that influence medication selection and cardiovascular risk stratification throughout hospitalization. Still, like weight disorders, it often serves as a secondary diagnosis (89.6\% of the cases in our dataset). The model detected 61.3\% of 23,510 human-annotated cases at the standard decision boundary, rising to 86.0\% at the value of 0.1. The model identified 25,529 additional potential hypertension cases in the test set that were not coded. Manual review of 50 randomly sampled cases revealed that 74\% contained explicit mentions of hypertension that human coders missed, and 10\% contained patients on antihypertensive medications without explicit diagnosis documentation. In the remaining 16\% the model inferred hypertension from related cardiovascular conditions or symptoms.

\subsection{Documentation quality affects model performance}

Beyond systematic under-coding, we identified two documentation-related challenges that impact model performance: ambiguous documentation leading to imprecise coding, and missing documentation despite correct coding.

\subsubsection{Imprecise coding due to poor documentation visibility}

The ICD-10 system comprises several codes for diabetes mellitus. Danish hospitals receive the same reimbursement regardless of the diabetes mellitus code. Alternatives include the unspecific diabetes code (E14) and the specific ones, E10 and E11. Among 926 unspecified diabetes cases in the test set, the model agreed with human coding in only 314 cases. In 31\% of unspecified diabetes annotations, the model showed high confidence ($>$0.1) for either type 1 or type 2 while showing minimal confidence ($<$0.1) for the unspecific code. To understand this disagreement, we randomly sampled 50 cases where the model predicted specific types of diabetes, while humans coded unspecified diabetes. Using our explanation method in a manual analysis, we found that 74\% (37 cases) contained explicit mentions of either type 1 or type 2 diabetes. In the remaining cases, the model inferred type 2 diabetes from metformin prescriptions.

To understand why unspecified diabetes was coded rather than type 1 or type 2 diabetes, we examined the full clinical notes. We discovered that while physicians had documented the specific type of diabetes (e.g., ``type 2 diabetes''), it was typically mentioned only once and buried deep within long notes. Meanwhile, the generic term ``diabetes'' appeared repeatedly throughout the same narratives. This pattern suggests an alternative explanation for unspecific coding. When secretaries code these cases, the secretary must read the entire dense clinical documentation. The single mention of diabetes type, buried within excessive amounts of generic diabetes references, is easily missed. These findings suggest that poorly written documentation significantly contributes to imprecise coding.

\subsubsection{Missing documentation with correct coding}

We also discovered cases where conditions were coded without the physician documenting them in the clinical notes. Among 7,454 annotated heart failure cases, the model predicted only 67.4\%. In 1,312 cases (17.6\%), the model's confidence was below 0.2, indicating uncertainty about the heart failure codes. Random sampling of 50 low-confidence cases revealed clear evidence of heart failure in only 38\% (19 cases), and half of these relied on evidence such as a low ejection fraction rather than explicit mentions of heart failure. There was no clear evidence for heart failure in the remaining 62\% of the cases.

Our explanation method revealed that when coding systematically was associated with incomplete documentation, the model learned to use administrative information as predictive features. The model would predict codes within that specialty based solely on the administrative information in the note, without any supporting clinical evidence. This creates a problematic bias where the model tends to learn documentation habits rather than general clinical patterns.

These findings highlight how documentation quality creates distinct challenges: crucial points are obscured by excessive information, leading to imprecise coding, while the absence of documentation forces models to rely on proxy correlations. Both patterns compromise the reliability of automated coding systems, suggesting that improving documentation practices may be as crucial as addressing reimbursement incentives.

\section{Discussion}

\subsection{What causes the under-coding of secondary diagnoses?}

Our analysis suggests systematic undercoding of secondary diagnoses in Denmark during the study period. Reimbursement models could drive this. While the WHO's ICD-10 guidelines mandate comprehensive coding of all conditions affecting patient management, the Danish reimbursement system may create opposing incentives~\citep{InternationalStatisticalClassification2004}. Hospitals receive payment primarily based on primary diagnoses and procedures. Only under certain conditions will some secondary diagnoses increase reimbursement, and the XX chapter (V00--Y99), which includes suicide-related codes, is excluded entirely from reimbursement calculations~\citep{LogiktabellerLPR}.

This incentive structure produces stark differences in coding practices. Emergency departments at Beth Israel Deaconess Medical Center in the US reported a median of fourteen ICD-10 codes per patient, compared to just two in Danish emergency departments~\citep{edinAutomatedMedicalCoding2023a,johnsonMIMICIVFreelyAccessible2023}. US hospitals usually employ dedicated coding staff and are reimbursed for all diagnoses. In contrast, Danish hospitals relied on medical secretaries until 2016, after which physicians took over handling coding alongside their primary responsibilities. With time pressures and minimal financial incentive to document secondary conditions, comprehensive coding becomes impractical.

\subsection{Implications of poor coding quality for automated coding systems}

Our analysis reveals a fundamental paradox in how models learn from incompletely coded data. Through detailed case studies of suicide-related behaviors, weight disorders, and hypertension, we discovered that the model correctly identified relevant clinical features for these conditions yet assigned them low confidence scores. This indicates that the model has learned two conflicting signals: how to recognize these conditions from clinical text, and that these recognized patterns should receive low confidence scores because they often go uncoded in the training data.

By experimenting with lower decision boundaries, we showed that the model's poor performance could be corrected. For suicide-related codes, lowering the boundary from 0.5 to 0.05 improved detection from 21.3\% to 64.5\%. Similar improvements occurred for weight disorders (52.8\% to 78.6\% at a 0.1 boundary) and hypertension (61.3\% to 86.0\% at a 0.1 boundary). Manual validation of model-identified cases that lacked human annotations showed accuracies of 76--86\%, confirming that these were genuine missed diagnoses rather than model errors. The model's low confidence scores reflected not its diagnostic capability, but rather the systematic under-coding in its training and test data.

These findings suggest that systematically selecting optimal decision boundaries for each code could substantially improve model performance. However, implementing this solution presents significant challenges. Determining optimal boundaries requires reliable ground truth data, which is precisely what we lack. While one could manually tune boundaries for each code based on validation studies like ours, this approach becomes impractical when scaled to thousands of ICD-10 codes. This process would need to be repeated whenever models are retrained. Synthetic data generation offers a potential solution, where artificial examples could provide the clean data for calibrating the decision boundaries for each code. Alternatively, one might use expert-curated subsets of data for boundary tuning or machine-learning frameworks that rely less heavily on human annotations~\citep{motzfeldtCodeHumansMultiAgent2025}.

\subsection{Can the model improve the coding quality?}

Despite these challenges, our results demonstrate that the model could meaningfully improve current coding workflows. When evaluated against human annotations, the model correctly predicts all assigned codes for 54.6\% of cases, enabling complete automation for half of the coding workload (with human review). For the remaining cases, 95.5\% of the human-annotated codes appear within the model's top 10 predictions. This means that in most cases where full automation is not possible, coders could find the correct codes by reviewing a short list of suggestions rather than searching through thousands of possibilities. Only in rare instances would the model fail to highlight the relevant code among its top suggestions, requiring traditional manual coding.

More importantly, such a workflow could address the systematic under-coding of secondary diagnoses. By surfacing relevant codes in its suggestions, the model reduces the burden of searching through entire clinical notes. As shown in Figure~\ref{fig:recall}b, 90\% of the human-coded secondary diagnoses appear among the model's top 10 predictions. Such human-AI collaboration could enhance coding completeness while preserving the clinical judgment necessary for complex cases.

While these results are promising, the model's actual effectiveness requires validation through user studies in clinical settings. Such studies would determine whether the model improves coding speed and completeness in practice, or whether it merely adds to coders' cognitive burden.

\subsection{Challenge of class imbalance}

Medical coding exhibits extreme class imbalance, with many codes appearing rarely in clinical encounters. Our results show a strong correlation between code frequency and human-model agreement, with rare codes typically achieving lower F1 scores. Traditional approaches to address class imbalance face unique challenges in the multilabel setting of medical coding. Upsampling rare codes is difficult because they frequently co-occur with common codes in the same cases. Amplifying examples containing rare codes would simultaneously amplify the common codes they appear with, limiting the effectiveness of this approach.

Other methods, such as focal loss, have also shown limited success in medical coding~\citep{liuEffectiveConvolutionalAttention2021}. These methods rely on the assumption that training labels are correct. However, if rare codes are systematically under-coded, upweighting them would amplify incorrect patterns in the training data.

The intersection of class imbalance and label noise presents a unique challenge. Future work should explore methods that jointly address both issues, potentially through semi-supervised approaches that leverage the model's demonstrated ability to identify missed diagnoses. Synthetic data generation could provide clean training signals for rare codes, while techniques for learning from noisy labels could help the model distinguish between actual patterns and annotation artifacts.

\section{Methods}

\subsection{Patient cohorts}

We used all electronic health records (EHR) from hospitals in the Capital Region of Denmark and Region Zealand, spanning the period from 2006 to 2016. In total, there are approximately 16 million electronic health records from 2.6 million patients~\citep{musePopulationwideAnalysisHospital2023,henriksenGenomewideAssociationsSpanning2025}. From the EHRs, we used the clinical notes, laboratory results, prescribed and administered medications, and the level-3 and level-5 ICD-10 codes.

\begin{figure}[h]
\centering
\includegraphics[width=0.6\textwidth]{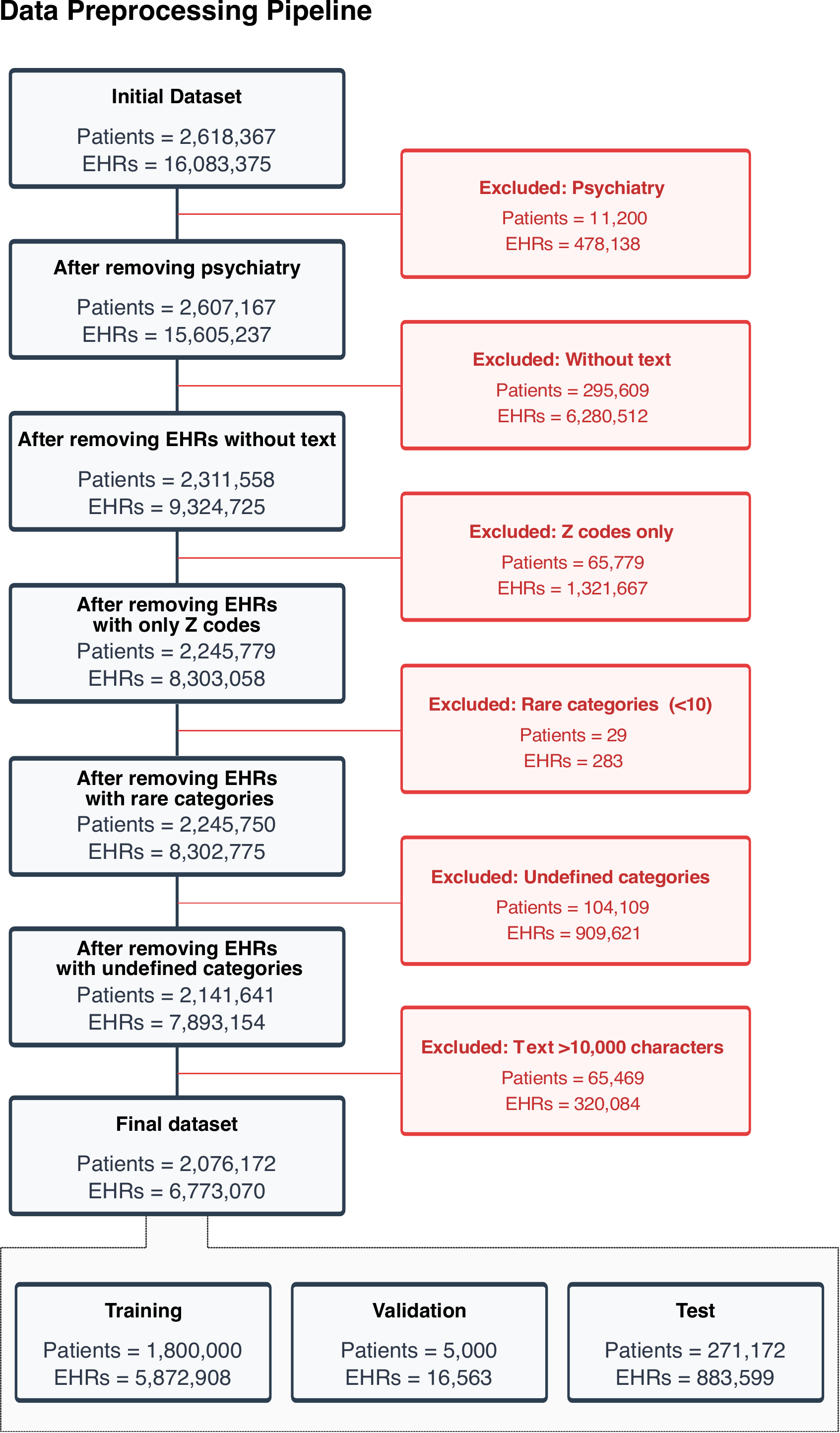}
\caption{The data preprocessing pipeline for the study.}
\label{fig:pipeline}
\end{figure}

\subsubsection{Data preprocessing}

One electronic health record defines one patient's course. Each patient course comprises several patient encounters and may include various types of clinical notes, medications, and laboratory results. We concatenate all into one string per patient course. Furthermore, we concatenate all the ICD-10 codes and remove duplicate codes.

Figure~\ref{fig:pipeline} depicts the data preprocessing pipeline. We first removed all cases lacking medical code annotations, as these provided no supervisory signal for training. We also excluded cases without discharge summaries, as preliminary analysis revealed that discharge summaries contained the most diagnostically relevant information. By removing cases lacking discharge summaries, we could train the model on a smaller, more informative dataset without incurring performance loss while achieving faster training times.

We removed cases containing only Z codes, which primarily represent administrative encounters such as vaccinations or no-shows that provide limited diagnostic information. We also excluded cases with categories (level-3 codes) appearing fewer than 10 times across the entire dataset to ensure sufficient training examples for each diagnostic category.

Cases with categories not defined in the Danish modification of the ICD-10 code system were removed to eliminate annotation errors, such as typos or incorrect use of the code system.

To manage computational resources and prevent extremely long sequences, we excluded cases where the combined clinical notes exceeded 10,000 characters. Additionally, all psychiatry cases were removed as they typically contained substantially longer clinical documentation that exceeded our processing constraints. This was not the case for child and adolescent psychiatry, which we retained.

\subsection{Model}

We use the PLM-CA architecture, which comprises a language model and a label-wise attention layer~\citep{edinAutomatedMedicalCoding2023a,edinUnsupervisedApproachAchieve2024a,huangPLMICDAutomaticICD2022,mullenbachExplainablePredictionMedical2018}. The language model encodes all the input tokens in 128 token windows, without overlap, into token embeddings. The label-wise attention layer attends to all the token embeddings and outputs one confidence score for each ICD-10 code.

We use the code released by Edin et al.\ (2024) for training and evaluating the models~\citep{edinUnsupervisedApproachAchieve2024a}. We use a BERT model trained on Danish text as the language model~\citep{devlinBERTPretrainingDeep2019}. We feed raw text to the model and use the human-annotated ICD-10 codes as ground truth. We train the model using binary cross-entropy for ten epochs using a learning rate of $5 \times 10^{-5}$ with a linear learning rate scheduler and AdamW optimizer~\citep{kingmaAdamMethodStochastic2017}. We stop the training early if there are no improvements in the mean average precision on the validation set for two consecutive epochs. Because our dataset has fewer codes per example than MIMIC (2 vs 14), we used a larger batch size of 128---compared to 16 used by Edin et al.\ (2024)~\citep{edinUnsupervisedApproachAchieve2024a}---to ensure sufficient signal per batch. We truncated the inputs to a maximum of 10,000 tokens due to memory constraints. This was rarely necessary since we had already filtered texts longer than 10,000 characters.

\subsection{Evaluation}

Since PLM-CA outputs confidence scores rather than binary predictions, similar to Edin et al.\ (2023)~\citep{edinAutomatedMedicalCoding2023a}, we employ two complementary strategies for selecting and evaluating final code sets against human annotations: threshold-based and ranking-based metrics.

\subsubsection{Threshold-based metrics}

These metrics use a cutoff value (threshold) to convert confidence scores into binary predictions. For example, with a threshold of 0.5, any code with a confidence score above 0.5 is predicted as present, while codes below this value are rejected. We determine the optimal threshold by testing different values and selecting the one that maximizes the micro F1 score on our validation dataset.

F1 Scores balance precision (avoiding incorrect predictions) and recall (finding all human-annotated codes) through their harmonic mean. We compute both micro and macro variants:

The F1 micro counts all correct and incorrect predictions across the entire dataset before calculating performance. This approach gives more weight to frequently occurring codes, as they contribute more predictions to the total count.

The F1 macro calculates performance separately for each code and then averages these scores. This treats rare and common codes equally. When F1 micro exceeds F1 macro, it indicates the model performs better on frequent codes than rare ones.

The Exact Match Ratio provides the strictest performance measure by calculating the percentage of cases where the model's selected codes match the human-annotated codes perfectly, with no missing or extra codes. This metric directly indicates what fraction of cases could be fully automated without human review.

\subsubsection{Ranking-based metrics}

Rather than selecting codes above a threshold, these metrics assess how well the model ranks codes based on confidence.

Recall@K measures the percentage of human-annotated codes appearing within the model's top K most confident predictions. For instance, Recall@5 checks whether the human-annotated codes appear somewhere in the top 5 recommendations. We evaluate at $K \in \{5, 10, 15\}$ to determine if human coders, when reviewing the model's suggestions, can find the necessary codes without extensive searching.

However, Recall@K has an inherent limitation: when a case has more human-annotated codes than K (e.g., seven human-annotated codes but $K = 5$), achieving 100\% recall becomes impossible. Precision@Recall addresses this by automatically adjusting K to match the number of human-annotated codes for each case, then measuring precision at that cutoff.

Mean Average Precision (MAP) evaluates the entire ranking without any fixed cutoff. It examines the position of each human-annotated code in the ranked list and calculates precision at those positions. The metric rewards models that place human-annotated codes near the top of their rankings. A model that ranks all human-annotated codes in positions 1--3 scores higher than one ranking them in positions 10--12, even if both eventually include all human-annotated codes.

\subsubsection{Manual analysis}

\begin{figure}[th]
\centering
\includegraphics[width=\textwidth]{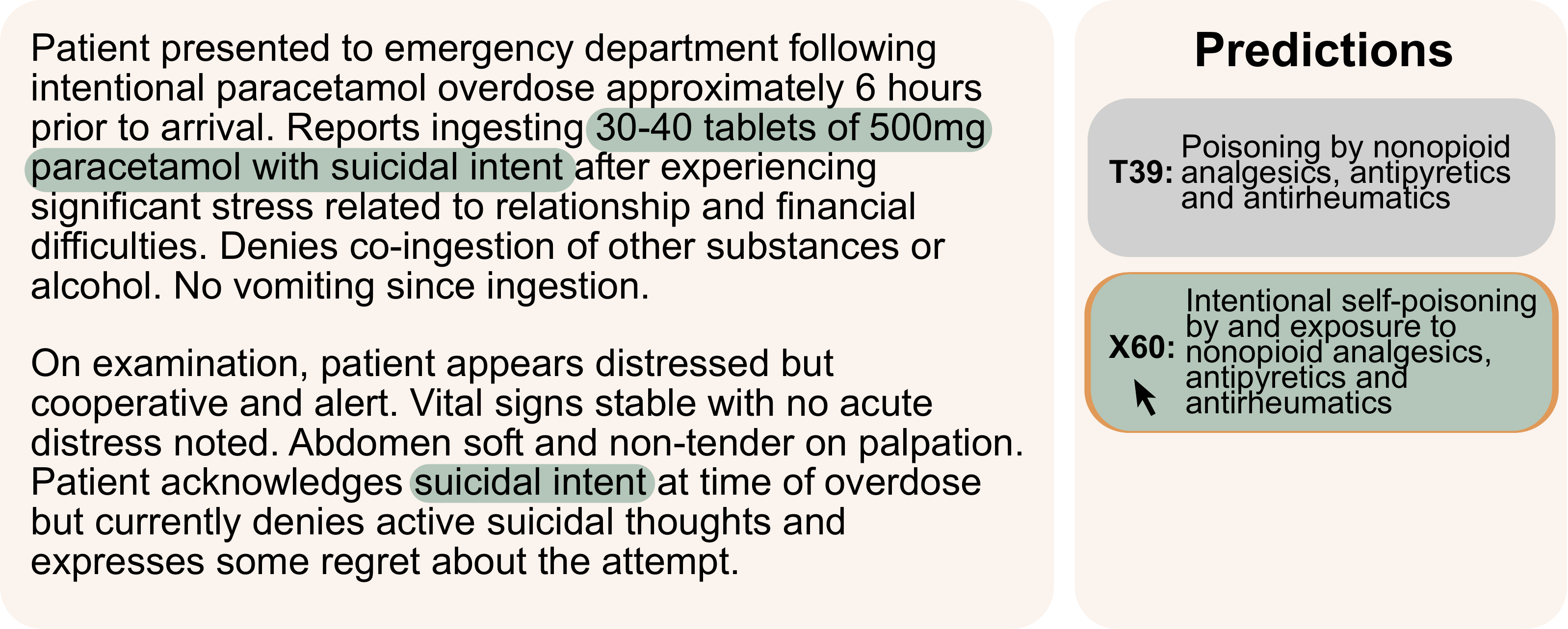}
\caption{The interface we used for our manual explainability analysis. The clinical note is shown to the left, and the model's predictions are to the right. When hovering the mouse over a code prediction, the AttInGrad explanations are highlighted in the text of the clinical note. By visualizing the factors that caused the model prediction, we could quickly identify the elements that led to the human-model disagreements. The text in this example is AI-generated and does not relate to the data used in the study.}
\label{fig:interface}
\end{figure}

To understand model behavior and error patterns, we developed an interactive interface for analyzing PLM-CA predictions (Figure~\ref{fig:interface}). The interface displays clinical information on the left side alongside the model's code predictions on the right. The interface enables exploration of the model's decision-making process through interactive highlighting. When hovering over any predicted code, the clinical text dynamically highlights the words that influenced the model's prediction for that specific code. We used AttInGrad (Attention Input Gradients), an explanation method, to identify which words to highlight~\citep{edinUnsupervisedApproachAchieve2024a}.

This visualization approach enables the rapid identification of patterns in both successful predictions and errors. By examining which text segments the model attends to, we can determine whether errors stem from focusing on irrelevant information, missing critical clinical context, or misinterpreting ambiguous terminology. The interface thus serves as a valuable tool for debugging.

\section{Conclusion}

This study represents the most extensive evaluation of automated medical coding to date, analyzing 1.8 million patients and 5.8 million electronic health records across all medical specialties in Eastern Denmark. Our findings reveal both the promise and challenges of deploying AI systems trained on real-world clinical data.

The model demonstrates strong performance when evaluated against human annotations, achieving sufficient accuracy to automate coding for 55.6\% of cases and providing valuable suggestions for the remaining 44.4\%. However, our analysis uncovered systematic under-coding of secondary diagnoses in the training data. When we investigated cases where the model disagreed with human coders, manual validation showed the model was correct 76--86\% of the time, identifying genuine conditions that humans had not coded.

Our findings have implications beyond Eastern Denmark. Healthcare systems worldwide face similar pressures: time constraints, reimbursement structures that prioritize primary diagnoses, and competing demands on healthcare professionals' attention. These factors shape coding practices in ways that compromise the quality of medical data, affecting research, public health surveillance, and resource allocation. The under-coding of suicide attempts exemplifies how incomplete documentation can hamper responses to public health crises.

Despite these challenges, automated coding systems offer a path forward. By surfacing commonly missed diagnoses, these tools can help healthcare professionals document the complete clinical picture without significant additional time investment. The success of automated medical coding ultimately depends on recognizing that these systems learn from imperfect data shaped by healthcare practices and policies. Only by improving both the technology and the underlying documentation practices can we enhance the quality and completeness of medical records that underpin modern healthcare.

\section*{Approvals}

This study was approved by The Danish Data Protection Agency (ref: 514--0255/18--3000, 514--0254/18--3000, SUND-2016--50), The Danish Health Data Authority (ref: FSEID-00003724 and FSEID-00003092), and The Danish Patient Safety Authority (3--3013-1731/1/). The study was approved as a registry study, for which consent and ethical approval are not required under Danish law. The study was carried out in accordance with relevant guidelines and regulations.

\section*{Data Availability}

The data that support the findings of this study are available from the Danish Health Data Authority and the Danish Patient Safety Authority. Application for data access can be made to the Danish Health Data Authority and the Danish Patient Safety Authority (contact: \texttt{servicedesk@sundhedsdata.dk}). Anyone wanting access to the data and to use them for research will be required to meet research credentialing requirements as outlined at the authority's web site: \url{https://sundhedsdatastyrelsen.dk/da/english/health_data_and_registers/research_services}.

\section*{Code Availability}

The code used for the model architecture, training, and evaluation is publicly available at \url{https://github.com/JoakimEdin/explainable-medical-coding}. The specific hyperparameters and training configurations used in this study are detailed in the Methods section. Custom scripts strictly related to the preprocessing of the sensitive registry data are not publicly available due to data protection regulations.

\section*{Acknowledgements}

This research was partially funded by the Innovation Fund Denmark via the Industrial Ph.D.\ Program (grant no.\ 2050-00040B) and via the Novo Nordisk Foundation (grant no.\ NNF14CC0001 and NNF24SA0098829). The funders played no role in study design, data collection, analysis and interpretation of data, or the writing of this manuscript. The authors acknowledge the Danish Health Data Authority and the Danish Patient Safety Authority for providing access to the registry data used in this study.

\section*{Author Contributions}

J.E.\ and S.B.\ conceived the study. J.E.\ developed the methodology and software, and performed the primary data analysis. S.B.\ and L.M.\ supervised the project. S.B., S.B.B., A.K.K., and I.L.\ assisted in analyzing the results. J.E.\ wrote the original draft. S.B., L.M., S.B.B., and A.K.K.\ contributed to the writing and editing of the manuscript. All authors critically revised and approved the final manuscript.

\section*{Competing Interests}

J.E.\ and L.M.\ are employees of Corti, a company that develops and sells medical coding software. L.M.\ additionally holds equity in the company. S.B.\ holds equity in Hoba Therapeutics, Novo Nordisk and Eli Lilly. All other authors declare no financial or non-financial competing interests.

\bibliographystyle{unsrtnat}
\bibliography{references}

@article{burnsSystematicReviewDischarge2012,
  title = {Systematic Review of Discharge Coding Accuracy},
  author = {Burns, E.M. and Rigby, E. and Mamidanna, R. and Bottle, A. and Aylin, P. and Ziprin, P. and Faiz, O.D.},
  year = 2012,
  month = mar,
  journal = {Journal of Public Health (Oxford, England)},
  volume = {34},
  number = {1},
  pages = {138--148},
  issn = {1741-3842},
  doi = {10.1093/pubmed/fdr054},
  urldate = {2022-05-24},
  pmcid = {PMC3285117},
  pmid = {21795302}
}

@article{changRegionalChangesCharcoalBurning2014,
  title = {Regional {{Changes}} in {{Charcoal-Burning Suicide Rates}} in {{East}}/{{Southeast Asia}} from 1995 to 2011: {{A Time Trend Analysis}}},
  shorttitle = {Regional {{Changes}} in {{Charcoal-Burning Suicide Rates}} in {{East}}/{{Southeast Asia}} from 1995 to 2011},
  author = {Chang, Shu-Sen and Chen, Ying-Yeh and Yip, Paul S. F. and Lee, Won Jin and Hagihara, Akihito and Gunnell, David},
  year = 2014,
  month = apr,
  journal = {PLOS Medicine},
  volume = {11},
  number = {4},
  pages = {e1001622},
  publisher = {Public Library of Science},
  issn = {1549-1676},
  doi = {10.1371/journal.pmed.1001622},
  urldate = {2025-07-07},
  langid = {english}
}

@article{cipollaImportanceComorbiditiesIschemic2018,
  title = {The Importance of Comorbidities in Ischemic Stroke: {{Impact}} of Hypertension on the Cerebral Circulation},
  shorttitle = {The Importance of Comorbidities in Ischemic Stroke},
  author = {Cipolla, Marilyn J and Liebeskind, David S and Chan, Siu-Lung},
  year = 2018,
  month = dec,
  journal = {Journal of Cerebral Blood Flow \& Metabolism},
  volume = {38},
  number = {12},
  pages = {2129--2149},
  issn = {0271-678X},
  doi = {10.1177/0271678X18800589},
  urldate = {2025-07-08},
  pmcid = {PMC6282213},
  pmid = {30198826}
}

@article{devlinBERTPretrainingDeep2019,
  title = {{{BERT}}: {{Pre-training}} of {{Deep Bidirectional Transformers}} for {{Language Understanding}}},
  shorttitle = {{{BERT}}},
  author = {Devlin, Jacob and Chang, Ming-Wei and Lee, Kenton and Toutanova, Kristina},
  year = 2019,
  month = may,
  journal = {arXiv:1810.04805 [cs]},
  eprint = {1810.04805},
  primaryclass = {cs},
  urldate = {2021-10-04},
  archiveprefix = {arXiv}
}

@article{dongAutomatedClinicalCoding2022,
  title = {Automated Clinical Coding: What, Why, and Where We Are?},
  shorttitle = {Automated Clinical Coding},
  author = {Dong, Hang and Falis, Mat{\'u}{\v s} and Whiteley, William and Alex, Beatrice and Matterson, Joshua and Ji, Shaoxiong and Chen, Jiaoyan and Wu, Honghan},
  year = 2022,
  month = oct,
  journal = {npj Digital Medicine},
  volume = {5},
  number = {1},
  pages = {1--8},
  publisher = {Nature Publishing Group},
  issn = {2398-6352},
  doi = {10.1038/s41746-022-00705-7},
  urldate = {2023-01-11},
  copyright = {2022 The Author(s)},
  langid = {english}
}

@article{edinAutomatedMedicalCoding2023a,
  title = {Automated {{Medical Coding}} on {{MIMIC-III}} and {{MIMIC-IV}}: {{A Critical Review}} and {{Replicability Study}}},
  shorttitle = {Automated {{Medical Coding}} on {{MIMIC-III}} and {{MIMIC-IV}}},
  author = {Edin, Joakim and Junge, Alexander and Havtorn, Jakob Drachmann and Borgholt, Lasse and Maistro, Maria and Ruotsalo, Tuukka and Maal{\o}e, Lars},
  year = 2023,
  journal = {Annual International ACM SIGIR Conference on Research and Development in Information Retrieval},
  doi = {10.48550/arXiv.2312.13533},
  urldate = {2024-02-07},
  langid = {english}
}

@inproceedings{edinUnsupervisedApproachAchieve2024a,
  title = {An {{Unsupervised Approach}} to {{Achieve Supervised-Level Explainability}} in {{Healthcare Records}}},
  booktitle = {Proceedings of the 2024 {{Conference}} on {{Empirical Methods}} in {{Natural Language Processing}}},
  author = {Edin, Joakim and Maistro, Maria and Maal{\o}e, Lars and Borgholt, Lasse and Havtorn, Jakob Drachmann and Ruotsalo, Tuukka},
  editor = {{Al-Onaizan}, Yaser and Bansal, Mohit and Chen, Yun-Nung},
  year = 2024,
  month = nov,
  pages = {4869--4890},
  publisher = {Association for Computational Linguistics},
  address = {Miami, Florida, USA},
  doi = {10.18653/v1/2024.emnlp-main.280},
  urldate = {2025-03-10}
}

@article{hawtonLongTermEffect2013,
  title = {Long Term Effect of Reduced Pack Sizes of Paracetamol on Poisoning Deaths and Liver Transplant Activity in {{England}} and {{Wales}}: Interrupted Time Series Analyses},
  shorttitle = {Long Term Effect of Reduced Pack Sizes of Paracetamol on Poisoning Deaths and Liver Transplant Activity in {{England}} and {{Wales}}},
  author = {Hawton, Keith and Bergen, Helen and Simkin, Sue and Dodd, Sue and Pocock, Phil and Bernal, William and Gunnell, David and Kapur, Navneet},
  year = 2013,
  month = feb,
  journal = {BMJ},
  volume = {346},
  pages = {f403},
  publisher = {British Medical Journal Publishing Group},
  issn = {1756-1833},
  doi = {10.1136/bmj.f403},
  urldate = {2025-07-18},
  chapter = {Research},
  copyright = {\copyright{} Hawton et al 2013. This is an open-access article distributed under the terms of the Creative Commons Attribution Non-commercial License, which permits use, distribution, and reproduction in any medium, provided the original work is properly cited, the use is non commercial and is otherwise in compliance with the license. See: http://creativecommons.org/licenses/by-nc/2.0/  and  http://creativecommons.org/licenses/by-nc/2.0/legalcode.},
  langid = {english},
  pmid = {23393081}
}

@article{henriksenGenomewideAssociationsSpanning2025,
  title = {Genome-Wide Associations Spanning 194 in-Hospital Drug Dosage Change Phenotypes Highlight Diverse Genetic Backgrounds in Concurrent Drug Therapy},
  author = {Henriksen, Alexander Pil and Rodr{\'i}guez, Cristina Leal and Currant, Hannah and Louloudis, Ioannis and Biel, Jorge Hernansanz and {Herrero-Zazo}, Maria and Birney, Ewan and Hansen, Thomas Folkmann and Mazzoni, Gianluca and Haue, Amalie Dahl and Bundgaard, Henning and Erikstrup, Christian and Dinh, Khoa Manh and Quinn, Liam and Bruun, Mie Topholm and Hjalgrim, Henrik and S{\o}rensen, Erik and Mikkelsen, Christina and Schwinn, Michael and Pedersen, Ole Birger Vestager and Ullum, Henrik and Ostrowski, Sisse Rye and Banasik, Karina and Brunak, S{\o}ren},
  year = 2025,
  month = jan,
  journal = {Computational and Structural Biotechnology Journal},
  volume = {28},
  pages = {239--248},
  publisher = {Elsevier},
  issn = {2001-0370},
  doi = {10.1016/j.csbj.2025.06.042},
  urldate = {2025-09-16},
  langid = {english},
  pmid = {40678446}
}

@inproceedings{huangPLMICDAutomaticICD2022,
  title = {{{PLM-ICD}}: {{Automatic ICD Coding}} with {{Pretrained Language Models}}},
  shorttitle = {{{PLM-ICD}}},
  booktitle = {Proceedings of the 4th {{Clinical Natural Language Processing Workshop}}},
  author = {Huang, Chao-Wei and Tsai, Shang-Chi and Chen, Yun-Nung},
  year = 2022,
  month = jul,
  pages = {10--20},
  publisher = {Association for Computational Linguistics},
  address = {Seattle, WA},
  doi = {10.18653/v1/2022.clinicalnlp-1.2},
  urldate = {2022-10-12}
}

@incollection{InternationalStatisticalClassification2004,
  title = {International Statistical Classification of Diseases and Related Health Problems. 1},
  year = 2004,
  edition = {2nd. ed},
  address = {Geneva},
  isbn = {978-92-4-154649-2},
  langid = {english}
}

@misc{jiUnifiedReviewDeep2023,
  title = {A {{Unified Review}} of {{Deep Learning}} for {{Automated Medical Coding}}},
  author = {Ji, Shaoxiong and Sun, Wei and Li, Xiaobo and Dong, Hang and Taalas, Ara and Zhang, Yijia and Wu, Honghan and Pitk{\"a}nen, Esa and Marttinen, Pekka},
  year = 2023,
  month = apr,
  number = {arXiv:2201.02797},
  eprint = {2201.02797},
  primaryclass = {cs},
  publisher = {arXiv},
  urldate = {2023-09-25},
  archiveprefix = {arXiv}
}

@article{johnsonMIMICIIIFreelyAccessible2016,
  title = {{{MIMIC-III}}, a Freely Accessible Critical Care Database},
  author = {Johnson, Alistair E. W. and Pollard, Tom J. and Shen, Lu and Lehman, Li-wei H. and Feng, Mengling and Ghassemi, Mohammad and Moody, Benjamin and Szolovits, Peter and Anthony Celi, Leo and Mark, Roger G.},
  year = 2016,
  month = may,
  journal = {Scientific Data},
  volume = {3},
  number = {1},
  pages = {160035},
  publisher = {Nature Publishing Group},
  issn = {2052-4463},
  doi = {10.1038/sdata.2016.35},
  urldate = {2022-04-19},
  copyright = {2016 The Author(s)},
  langid = {english}
}

@article{johnsonMIMICIVFreelyAccessible2023,
  title = {{{MIMIC-IV}}, a Freely Accessible Electronic Health Record Dataset},
  author = {Johnson, Alistair E. W. and Bulgarelli, Lucas and Shen, Lu and Gayles, Alvin and Shammout, Ayad and Horng, Steven and Pollard, Tom J. and Moody, Benjamin and Gow, Brian and Lehman, Li-wei H. and Celi, Leo A. and Mark, Roger G.},
  year = 2023,
  month = jan,
  journal = {Scientific Data},
  volume = {10},
  number = {1},
  pages = {1},
  publisher = {Nature Publishing Group},
  issn = {2052-4463},
  doi = {10.1038/s41597-022-01899-x},
  urldate = {2023-01-30},
  copyright = {2023 The Author(s)},
  langid = {english}
}

@misc{kingmaAdamMethodStochastic2017,
  title = {Adam: {{A Method}} for {{Stochastic Optimization}}},
  shorttitle = {Adam},
  author = {Kingma, Diederik P. and Ba, Jimmy},
  year = 2017,
  month = jan,
  number = {arXiv:1412.6980},
  eprint = {1412.6980},
  primaryclass = {cs},
  publisher = {arXiv},
  doi = {10.48550/arXiv.1412.6980},
  urldate = {2022-11-28},
  archiveprefix = {arXiv}
}

@article{leafMentalHealthService1996,
  title = {Mental Health Service Use in the Community and Schools: Results from the Four-Community {{MECA Study}}. {{Methods}} for the {{Epidemiology}} of {{Child}} and {{Adolescent Mental Disorders Study}}},
  shorttitle = {Mental Health Service Use in the Community and Schools},
  author = {Leaf, P. J. and Alegria, M. and Cohen, P. and Goodman, S. H. and Horwitz, S. M. and Hoven, C. W. and Narrow, W. E. and {Vaden-Kiernan}, M. and Regier, D. A.},
  year = 1996,
  month = jul,
  journal = {Journal of the American Academy of Child and Adolescent Psychiatry},
  volume = {35},
  number = {7},
  pages = {889--897},
  issn = {0890-8567},
  doi = {10.1097/00004583-199607000-00014},
  langid = {english},
  pmid = {8768348}
}

@incollection{limObesityComorbidConditions2025,
  title = {Obesity and {{Comorbid Conditions}}},
  booktitle = {{{StatPearls}}},
  author = {Lim, Yizhe and Boster, Joshua},
  year = 2025,
  publisher = {StatPearls Publishing},
  address = {Treasure Island (FL)},
  urldate = {2025-07-08},
  copyright = {Copyright \copyright{} 2025, StatPearls Publishing LLC.},
  langid = {english},
  lccn = {NBK574535},
  pmid = {34662049}
}

@inproceedings{liuEffectiveConvolutionalAttention2021,
  title = {Effective {{Convolutional Attention Network}} for {{Multi-label Clinical Document Classification}}},
  booktitle = {Proceedings of the 2021 {{Conference}} on {{Empirical Methods}} in {{Natural Language Processing}}},
  author = {Liu, Yang and Cheng, Hua and Klopfer, Russell and Gormley, Matthew R. and Schaaf, Thomas},
  year = 2021,
  month = nov,
  pages = {5941--5953},
  publisher = {Association for Computational Linguistics},
  address = {Online and Punta Cana, Dominican Republic},
  doi = {10.18653/v1/2021.emnlp-main.481},
  urldate = {2022-08-04}
}

@misc{LogiktabellerLPR,
  title = {{Logiktabeller for LPR}},
  urldate = {2025-07-19},
  howpublished = {https://sundhedsdatastyrelsen.dk/data-og-registre/sundhedsoekonomi/drg/drg-gruppering/logiktabeller-for-lpr},
  langid = {danish}
}

@article{macraeComorbidityChronicKidney2021,
  title = {Comorbidity in Chronic Kidney Disease: A Large Cross-Sectional Study of Prevalence in {{Scottish}} Primary Care},
  shorttitle = {Comorbidity in Chronic Kidney Disease},
  author = {MacRae, Clare and Mercer, Stewart W and Guthrie, Bruce and Henderson, David},
  year = 2021,
  month = feb,
  journal = {The British Journal of General Practice},
  volume = {71},
  number = {704},
  pages = {e243-e249},
  issn = {0960-1643},
  doi = {10.3399/bjgp20X714125},
  urldate = {2025-07-08},
  pmcid = {PMC7888754},
  pmid = {33558333}
}

@article{mannSuicidePreventionStrategies2005,
  title = {Suicide Prevention Strategies: A Systematic Review},
  shorttitle = {Suicide Prevention Strategies},
  author = {Mann, J. John and Apter, Alan and Bertolote, Jose and Beautrais, Annette and Currier, Dianne and Haas, Ann and Hegerl, Ulrich and Lonnqvist, Jouko and Malone, Kevin and Marusic, Andrej and Mehlum, Lars and Patton, George and Phillips, Michael and Rutz, Wolfgang and Rihmer, Zoltan and Schmidtke, Armin and Shaffer, David and Silverman, Morton and Takahashi, Yoshitomo and Varnik, Airi and Wasserman, Danuta and Yip, Paul and Hendin, Herbert},
  year = 2005,
  month = oct,
  journal = {JAMA},
  volume = {294},
  number = {16},
  pages = {2064--2074},
  issn = {1538-3598},
  doi = {10.1001/jama.294.16.2064},
  langid = {english},
  pmid = {16249421}
}

@article{matsubayashiEffectPublicAwareness2014,
  title = {The Effect of Public Awareness Campaigns on Suicides: {{Evidence}} from {{Nagoya}}, {{Japan}}},
  shorttitle = {The Effect of Public Awareness Campaigns on Suicides},
  author = {Matsubayashi, Tetsuya and Ueda, Michiko and Sawada, Yasuyuki},
  year = 2014,
  month = jan,
  journal = {Journal of Affective Disorders},
  volume = {152--154},
  pages = {526--529},
  issn = {0165-0327},
  doi = {10.1016/j.jad.2013.09.007},
  urldate = {2025-07-18}
}

@inproceedings{motzfeldtCodeHumansMultiAgent2025,
  title = {Code {{Like Humans}}: {{A Multi-Agent Solution}} for {{Medical Coding}}},
  booktitle = {Findings of the {{Association}} for {{Computational Linguistics}}: {{EMNLP}} 2025},
  author = {Motzfeldt, Andreas Geert and Edin, Joakim and Maal{\o}e, Lars and Rogers, Anna},
  year = 2025,
  volume = {Findings of the Association for Computational Linguistics: EMNLP 2025},
  publisher = {Association for Computational Linguistics}
}

@inproceedings{mullenbachExplainablePredictionMedical2018,
  title = {Explainable {{Prediction}} of {{Medical Codes}} from {{Clinical Text}}},
  booktitle = {Proceedings of the 2018 {{Conference}} of the {{North American Chapter}} of the {{Association}} for {{Computational Linguistics}}: {{Human Language Technologies}}, {{Volume}} 1 ({{Long Papers}})},
  author = {Mullenbach, James and Wiegreffe, Sarah and Duke, Jon and Sun, Jimeng and Eisenstein, Jacob},
  year = 2018,
  month = jun,
  pages = {1101--1111},
  publisher = {Association for Computational Linguistics},
  address = {New Orleans, Louisiana},
  doi = {10.18653/v1/N18-1100},
  urldate = {2023-02-07}
}

@article{musePopulationwideAnalysisHospital2023,
  title = {Population-Wide Analysis of Hospital Laboratory Tests to Assess Seasonal Variation and Temporal Reference Interval Modification},
  author = {Muse, Victorine P. and {Aguayo-Orozco}, Alejandro and Balaganeshan, Sedrah B. and Brunak, S{\o}ren},
  year = 2023,
  month = aug,
  journal = {Patterns},
  volume = {4},
  number = {8},
  publisher = {Elsevier},
  issn = {2666-3899},
  doi = {10.1016/j.patter.2023.100778},
  urldate = {2025-09-16},
  langid = {english}
}

@article{reutermorthorstIncidenceRatesDeliberate2016,
  title = {Incidence {{Rates}} of {{Deliberate Self-Harm}} in {{Denmark}} 1994--2011},
  author = {Reuter Morthorst, Britt and Soegaard, Bodil and Nordentoft, Merete and Erlangsen, Annette},
  year = 2016,
  journal = {Crisis},
  volume = {37},
  number = {4},
  pages = {256--264},
  issn = {0227-5910},
  doi = {10.1027/0227-5910/a000391},
  urldate = {2025-09-15},
  pmcid = {PMC5137321},
  pmid = {27278571}
}

@article{ryanClinicalPictureMajor1987,
  title = {The Clinical Picture of Major Depression in Children and Adolescents},
  author = {Ryan, N. D. and {Puig-Antich}, J. and Ambrosini, P. and Rabinovich, H. and Robinson, D. and Nelson, B. and Iyengar, S. and Twomey, J.},
  year = 1987,
  month = oct,
  journal = {Archives of General Psychiatry},
  volume = {44},
  number = {10},
  pages = {854--861},
  issn = {0003-990X},
  doi = {10.1001/archpsyc.1987.01800220016003},
  langid = {english},
  pmid = {3662742}
}

@article{ryanDiagnosingPediatricDepression2001,
  title = {Diagnosing Pediatric Depression},
  author = {Ryan, N. D.},
  year = 2001,
  month = jun,
  journal = {Biological Psychiatry},
  volume = {49},
  number = {12},
  pages = {1050--1054},
  issn = {0006-3223},
  doi = {10.1016/s0006-3223(01)01143-x},
  langid = {english},
  pmid = {11430846}
}

@article{santiagoImpactDiseaseComorbidities2021,
  title = {The {{Impact}} of {{Disease Comorbidities}} in {{Alzheimer}}'s {{Disease}}},
  author = {Santiago, Jose A. and Potashkin, Judith A.},
  year = 2021,
  month = feb,
  journal = {Frontiers in Aging Neuroscience},
  volume = {13},
  pages = {631770},
  issn = {1663-4365},
  doi = {10.3389/fnagi.2021.631770},
  urldate = {2025-07-08},
  pmcid = {PMC7906983},
  pmid = {33643025}
}

@article{stanfillPreparingICD10CMPCS2014,
  title = {Preparing for {{ICD-10-CM}}/{{PCS Implementation}}: {{Impact}} on {{Productivity}} and {{Quality}}},
  shorttitle = {Preparing for {{ICD-10-CM}}/{{PCS Implementation}}},
  author = {Stanfill, Mary H. and Hsieh, Kang Lin and Beal, Kathleen and Fenton, Susan H.},
  year = 2014,
  month = jul,
  journal = {Perspectives in Health Information Management},
  volume = {11},
  number = {Summer},
  pages = {1f},
  issn = {1559-4122},
  urldate = {2025-05-28},
  pmcid = {PMC4142514},
  pmid = {25214823}
}

@article{torokSystematicReviewMass2017,
  title = {A {{Systematic Review}} of {{Mass Media Campaigns}} for {{Suicide Prevention}}: {{Understanding Their Efficacy}} and the {{Mechanisms Needed}} for {{Successful Behavioral}} and {{Literacy Change}}},
  shorttitle = {A {{Systematic Review}} of {{Mass Media Campaigns}} for {{Suicide Prevention}}},
  author = {Torok, Michelle and Calear, Alison and Shand, Fiona and Christensen, Helen},
  year = 2017,
  journal = {Suicide and Life-Threatening Behavior},
  volume = {47},
  number = {6},
  pages = {672--687},
  issn = {1943-278X},
  doi = {10.1111/sltb.12324},
  urldate = {2025-07-18},
  langid = {english}
}

@article{tsengAdministrativeCostsAssociated2018,
  title = {Administrative {{Costs Associated With Physician Billing}} and {{Insurance-Related Activities}} at an {{Academic Health Care System}}},
  author = {Tseng, Phillip and Kaplan, Robert S. and Richman, Barak D. and Shah, Mahek A. and Schulman, Kevin A.},
  year = 2018,
  month = feb,
  journal = {JAMA},
  volume = {319},
  number = {7},
  pages = {691--697},
  issn = {0098-7484},
  doi = {10.1001/jama.2017.19148},
  urldate = {2023-02-06}
}

@book{worldhealthorganizationPreventingSuicideGlobal2014,
  title = {Preventing Suicide: A Global Imperative},
  shorttitle = {Preventing Suicide},
  author = {{World Health Organization}},
  year = 2014,
  publisher = {World Health Organization},
  address = {Geneva},
  urldate = {2025-07-04},
  chapter = {Japanese version published by Center for Suicide Prevention, National Institute of Mental Health, National Center of Neurology and Psychiatry},
  isbn = {978-92-4-156477-9},
  langid = {english}
}

@article{zuckerbrotImprovingRecognitionAdolescent2006,
  title = {Improving Recognition of Adolescent Depression in Primary Care},
  author = {Zuckerbrot, Rachel A. and Jensen, Peter S.},
  year = 2006,
  month = jul,
  journal = {Archives of Pediatrics \& Adolescent Medicine},
  volume = {160},
  number = {7},
  pages = {694--704},
  issn = {1072-4710},
  doi = {10.1001/archpedi.160.7.694},
  langid = {english},
  pmid = {16818834}
}

\end{document}